%% file: main.tex
\documentclass[11pt]{article}

\usepackage{acl}

\usepackage{times}
\usepackage{latexsym}
\usepackage[T1]{fontenc}
\usepackage[utf8]{inputenc}
\usepackage{microtype}
\usepackage{inconsolata}

\usepackage{graphicx}
\usepackage{amsmath,amssymb,mathtools}
\usepackage{amsthm}
\usepackage{booktabs}
\usepackage{multirow}

\theoremstyle{plain}

\theoremstyle{definition}

\theoremstyle{remark}

\title{Beyond the Stability-Exploration Dilemma: \\
Environmental Regularization for LLM Policy Optimization}

\author{
  \textbf{Xianlei Zhou}\textsuperscript{1,*},
  \textbf{Xiangdi Meng}\textsuperscript{1,*},
  \textbf{Yu He}\textsuperscript{1,2,\textdagger},
  \textbf{Tianyu Qi}\textsuperscript{3},
  \textbf{Shuyan Guan}\textsuperscript{1},
  \\
  \textbf{Xianli Zhang}\textsuperscript{1},
  \textbf{Jian Zhang}\textsuperscript{4},
  \textbf{Xin Li}\textsuperscript{1},
  \textbf{Qika Lin}\textsuperscript{5},
  \textbf{Jun Liu}\textsuperscript{2}
  \\[1ex]
  \textsuperscript{1}AMAP, Alibaba Group
  \quad
  \textsuperscript{2}Xi'an Jiaotong University
  \quad
  \textsuperscript{3}JD.com
  \\
  \textsuperscript{4}Beijing Normal University
  \quad
  \textsuperscript{5}National University of Singapore
  \\[0.5ex]
  \textsuperscript{*}Equal contribution.
  \quad
  \textsuperscript{\textdagger}Corresponding author.
  \\
  \texttt{\{zhouxianlei.zxl,gaolin.hy\}@alibaba-inc.com}
}

\begin{document}
\maketitle

\begin{abstract}
Policy optimization (PO) for Large Language Models faces a stability--exploration trade-off, currently mediated by an action-side Policy-KL regularizer. This puts practitioners in a double bind: keeping Policy-KL constrains response behavior and consumes the action-side exploration budget, while dropping it leaves the optimization without an explicit drift control. We argue for an alternative that breaks the dilemma by moving regularization to the input side. As training progresses, the distribution over training queries induced by the current policy drifts unchecked from its pre-RL reference distribution.Concretely, \textbf{Environment-Regularized Policy Optimization (ERPO)} introduces a \textbf{Query-KL (QKL)} term that bounds this query distribution shift, together with a dataset-static reference-derived per-query weight that biases each per-query update toward queries typical under the reference. The QKL gradient flows strictly through the query likelihood; the response score function used by policy-gradient estimators does not appear in the QKL term, so QKL exerts no direct gradient pressure on the response distribution---exploration is preserved. ERPO plugs into GRPO/PPO/REINFORCE-style pipelines without additional forward passes. On six mathematical reasoning benchmarks, \textbf{ERPO replaces} the standard Policy-KL regularizer while achieving effective control over query distribution drift, delivering stronger accuracy and substantially more stable behavior under high-temperature decoding and long-horizon training. Our source code are available at \url{https://github.com/AlibabaResearch/ERPO}
\end{abstract}

\input{tex/intro}
\input{tex/relatedworks}
\input{tex/preliminary_3}
\input{tex/method_2}
\input{tex/experiments}

\section{Conclusion}
By analyzing the coupling between the environment and the policy space in large language models, we decouple parameter regularization from the optimization objective during training. Specifically, we employ query-level KL divergence to bound the drift of the policy-induced query distribution $\rho_\theta$ from a pre-RL reference $\rho_{\theta_0}$. We further reweight the advantage by a dataset-static reference-derived per-query weight, biasing updates toward queries typical under $\rho_{\theta_0}$ and preventing premature convergence to suboptimal solutions. Experiments across multiple mathematical reasoning benchmarks demonstrate that the proposed ERPO method can achieve comparable KL divergence control without explicit policy regularization, while delivering superior performance. Furthermore, by sampling at different temperatures, we examine the evolution of sampling capability over long-term RL training, providing additional evidence of ERPO's stability during training.

\section*{Limitations}
Our experiments focus primarily on mathematical reasoning benchmarks and Qwen-family models, so the extent to which ERPO transfers to broader instruction-following, dialogue, code-generation, and multilingual settings remains to be validated. ERPO also relies on estimating query-level likelihoods or prevalence statistics during training; the quality and computational cost of these estimates may vary with the data-selection mechanism and model scale. Finally, we did not conduct an exhaustive sweep over the regularization coefficient, leaving more systematic hyperparameter analysis for future work.


\bibliography{example_paper}

\clearpage
\appendix
\input{tex/appendix}

\end{document}

%% file: tex/intro.tex
\section{Introduction}


Policy optimization (PO) methods have become the de facto recipe for post‑training large language models (LLMs), spanning trust‑region style updates (TRPO/PPO) and preference‑based objectives (DPO) together with broader RLHF/RLAIF variants~\citep{trpo15,ppo17,ouyang22,bai22,dpo23}. Despite impressive progress in mathematical reasoning and beyond, practitioners still face a persistent dilemma: \emph{how to trade off training stability against effective exploration}. In long‑horizon runs, optimization noise and distribution shift tend to accumulate, leading to oscillations and occasional collapses.

We argue that a key and under‑controlled source of instability is \emph{environment non‑stationarity} induced by the \textbf{query distribution}. Even with a fixed training corpus, the model's own sequence likelihood over training queries co‑evolves with the policy: as $\theta$ updates, the likelihood the model assigns to each prompt drifts, altering the effective training environment and amplifying gradient variance. This input‑side non‑stationarity mirrors classic RL settings in which either the initial‑state distribution or the transition kernel drifts over time; non‑stationary and robust RL therefore advocate explicit distributional control~\citep{padakandla21survey,iyengar05,nilim05}. A related lesson from imitation learning is that policy updates induce covariate (state) shift, motivating interactive data aggregation such as DAgger/AggreVaTe~\citep{ross11dagger,ross14aggrevate}.

\begin{figure}  
  \vspace{0pt} 
  \hspace{-18pt}
  \centering
  \includegraphics[width=0.8\columnwidth]{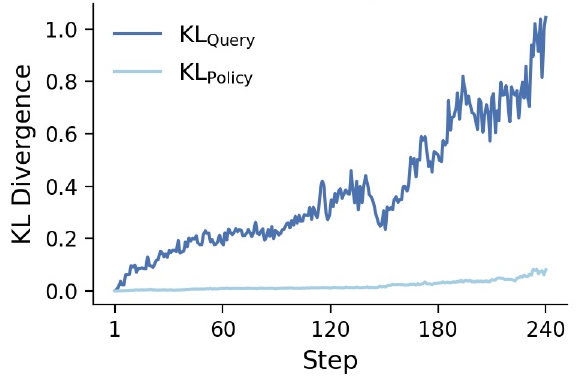}
  \caption{KL losses during GRPO training. The Query-KL (dark) rises while the Policy-KL (light) stays low, showing action-only KL does not stabilize the query process.}
  \label{fig:kl-loss-comparison}
  \vspace{-10pt}
\end{figure}

Recent LLM work formalizes prompt distributions (EVA, Align‑Pro~\citep{eva24,alignpro25}) or reweights training data (StablePrompt, WPO~\citep{stableprompt24,wpo24}), while mainstream RLHF PO focuses on action‑side Policy‑KL to an SFT reference~\citep{trpo15,ppo17,ouyang22}; neither directly constrains the query distribution. Empirically (Figure~\ref{fig:kl-loss-comparison}), under a fixed Policy‑KL budget the batch‑estimated Query‑KL rises steadily while Policy‑KL stays flat---the policy‑induced query distribution $\rho_\theta$ drifts unchecked from its pre‑RL reference $\rho_{\theta_0}$.

We treat the model's query likelihood as the input‑side statistic to regularize. We introduce \textbf{Query‑KL regularization (QKL)}, a penalty on the KL divergence between the current policy‑induced query distribution $\rho_\theta$ and a \emph{pre‑RL reference} $\rho_{\theta_0}$, limiting inter‑round drift of this likelihood while leaving the action space free to explore. In parallel, we propose a lightweight reference‑derived per‑query weight that biases each per‑query update toward queries typical under $\rho_{\theta_0}$, reducing estimator variance and improving robustness under high‑temperature decoding—where LLMs are especially sensitive to the long tail of decoding distributions~\citep{holtzman20,wang22sc}. Both components are estimator‑agnostic and compatible with GRPO/PPO/REINFORCE‑style implementations with minimal changes.
Figure~\ref{fig:method-overview} sketches ERPO: on top of GRPO we replace the usual Policy-KL with a Query-KL term, and during advantage computation we reweight per-query contributions by a reference-derived prior, yielding an environment-aware update while preserving action-side exploration.


We make four main contributions.
(1) \textbf{Query-environment control:} We treat the model's query likelihood as the regularizable input-side statistic, combining \emph{Query-KL (QKL)} to bound its drift from a pre-RL reference $\rho_{\theta_0}$ with a dataset-static reference-derived per-query weight to reduce variance and tame high-temperature behavior.
(2) \textbf{Estimator-agnostic instantiation:} The method adds a QKL term and a reference-derived per-query weight on top of GRPO/PPO/REINFORCE-style pipelines with minimal changes.
(3) \textbf{Stability evaluation:} We assess RL stability via \emph{multi-temperature} sampling paired with a \emph{multi-metric} suite (Pass@k, Pass@1, Avg@k), enabling comprehensive capability and robustness evaluation.
(4) \textbf{Empirical gains:} Across diverse reasoning benchmarks, the approach consistently improves accuracy.

%% file: tex/relatedworks.tex
\section{Related Works}
\label{sec:RW}

\subsection{Reinforcement Learning with Verifiable Rewards (RLVR)}
Reinforcement Learning with Verifiable Rewards represents a paradigm shift from traditional RLHF approaches by leveraging automatically verifiable outcomes rather than human preference annotations. This approach is particularly powerful for domains where ground truth can be objectively determined, such as mathematical reasoning, code generation, and logical problem solving. Models like AlphaCode~\citep{li2022competition} and recent mathematical reasoning~\citep{jeannotte2017conceptual,xia2025evaluating} systems leverage execution results and correctness verification as direct reward signals, eliminating the need for expensive human annotation.

\input{tex/figure2}

Process Reward Models (PRMs) have emerged as a sophisticated extension of RLVR, where intermediate steps in reasoning processes are evaluated and rewarded based on their correctness~\citep{uesato2022solving,lightman2023let}. Recent developments include tool-augmented reasoning systems~\citep{schick2023toolformer} and self-verification approaches~\citep{kojima2022large}, which combine language models with external verification tools to enable automatic reward computation for broader task domains.

While RLVR provides scalable and consistent training signals compared to subjective human preferences, it introduces unique challenges in handling high variance from sparse rewards and potential reward hacking behaviors. These stability issues motivate the need for robust training methodologies that can effectively leverage verifiable rewards while maintaining training stability.

\subsection{Reinforcement Learning Stability in Language Model Training}
The stability of reinforcement learning algorithms in language model training has become a critical research area due to unique challenges posed by discrete action spaces, large parameter spaces, and complex reward landscapes~\citep{sutton1998reinforcement}. Recent works have identified specific stability issues including reward hacking~\citep{gao2023scaling} and the alignment tax problem~\citep{dai2025mitigating}, where policy optimization can degrade downstream performance while improving target metrics.
Distribution shift during training has been recognized as a fundamental source of instability in policy gradient methods~\citep{reddy2020learning}. In language model contexts, this manifests as shifts in the query distribution during training, leading to high variance in gradient estimates and potential policy collapse~\citep{wen2024language}. 
Existing approaches primarily focus on action-space regularization through trust-region methods~\citep{schulman2015trust} and KL divergence penalties between the current policy and a reference policy~\citep{chegini2024salsasoupbasedalignmentlearning}.

Despite progress in understanding RL stability issues, there remains a notable gap in explicitly managing the input query distribution during training. Most current approaches focus on output regularization rather than addressing environmental shifts at the input level, leaving query distribution management as an underexplored avenue for improving training stability.

%% file: tex/figure2.tex

\begin{figure*}[t]
  \centering
  \includegraphics[width=0.8\textwidth]{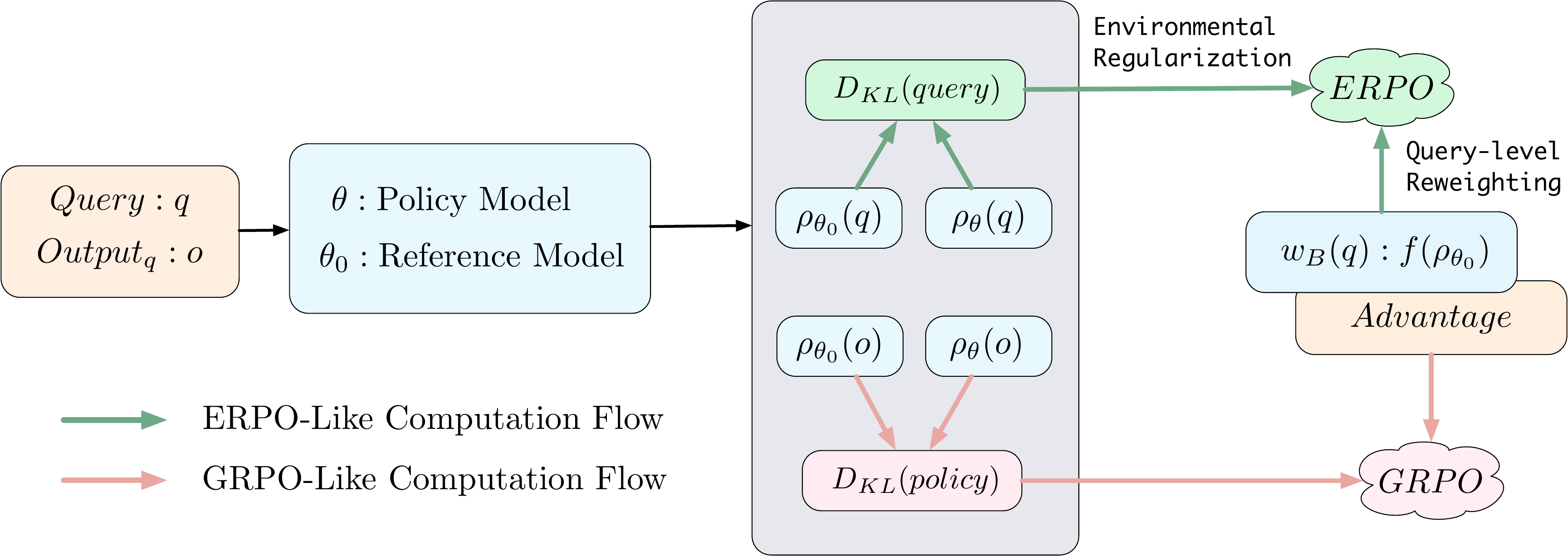}
  \caption{\textbf{Comparison of the GRPO and ERPO Computation Flows.}
  Given a query \(q\) and its output \(o\), the policy model \(\theta\) and the
  reference model \(\theta_{0}\) induce query distributions
  \(\rho_{\theta}(q)\) and \(\rho_{\theta_{0}}(q)\), as well as output
  distributions \(\rho_{\theta}(o)\) and \(\rho_{\theta_{0}}(o)\).
  The GRPO-like flow (red) applies policy-level KL regularization to the output
  distributions and optimizes the standard within-query advantage.
  In contrast, the ERPO-like flow (green) measures query-distribution drift
  through \(D_{\mathrm{KL}}(\text{query})\) for environmental regularization
  and reweights the advantage at the query level using
  \(w_B(q)=f(\rho_{\theta_{0}}(q))\)(Section \ref{sec:method-reweight-qkl}).}
  \label{fig:method-overview}
  \vspace{-5pt}
\end{figure*}

%% file: tex/preliminary_3.tex
\section{Preliminaries}
\label{sec:prelim}

\subsection{RLVR setting and notation}

We consider a standard generative--verify (RLVR) setting for training a large language model (LLM) with parameters $\theta$. Given a query $q\in\mathcal{Q}$, the LLM defines a response policy $\pi_\theta(o\mid q)$ over $o\in\mathcal{O}$, and a verifier returns a scalar reward $g(q,o)\in\mathbb{R}$. The underlying RL objective is
\begin{equation}
\label{eq:rl-objective}
J_{\mathrm{RL}}(\theta)
=
\mathbb{E}_{q\sim\rho_{train},\,o\sim\pi_\theta(\cdot\mid q)}
\big[g(q,o)\big],
\end{equation}
where $\rho_{\rm train}$ denotes the query distribution in the training set.


In practice, group-based or advantage-based variants of Eq.~\eqref{eq:rl-objective} are widely used (e.g., GRPO~\citep{Shao2024DeepSeekMath}, RLOO~\citep{ahmadian2024basicsrevisitingreinforcestyle}, DAPO~\citep{yu2025dapo}); our method is compatible with any such estimator. For concreteness in experiments we adopt the group-relative formulation of GRPO:
\begin{equation}
\label{eq:prelim-grpo-adv}
A^{\mathrm{GRPO}}_\theta\big(q,o^{(k)}\big)
=
\frac{g\big(q,o^{(k)}\big) - \mathrm{mean}(\mathbf{g})}{\mathrm{std}(\mathbf{g})},
\end{equation}
where $\mathbf{g}=\{g(q,o^{(k)})\}_{k=1}^{K}$ are the rewards of $K$ responses sampled from $\pi_\theta(\cdot\mid q)$ for the same query.

\subsection{Policy-induced query distribution and environment drift}
\label{sec:prelim_env_drift}

During LLM-RL training, the queries seen at each step are jointly shaped by the training corpus, curriculum design, difficulty filters, and active sampling schedulers. 
Rather than tying the analysis to such pipeline-specific factors, we
introduce a \emph{policy-only} notion: $\rho_\theta$ is the
\emph{policy-induced query distribution}, defined by
$\rho_\theta(q) = P_\theta(q)$ --- the model's autoregressive sequence
likelihood of $q$ treated as a self-terminating token sequence
(see Section~\ref{sec:method} for the autoregressive form). This model-induced
distribution is distinct from the exogenous $\rho_{\mathrm{train}}$, which
supplies queries to training batches and remains fixed in our experiments.

By construction, $\rho_\theta$ depends only on $\theta$, and the pre-RL reference $\rho_{\theta_0}$ inherits whatever data composition shaped $\pi_{\theta_0}$ (pretraining, SFT, or prior RL stages). As $\theta$ is updated, $\rho_\theta$ drifts away from $\rho_{\theta_0}$; without explicit constraint, this drift is unchecked. We call this \emph{policy-induced environment drift}, and quantify it by the KL divergence
\begin{equation}
\label{eq:envshift}
\begin{split}
\mathrm{EnvShift}(\theta)
&:= \mathrm{KL}\!\left(\rho_\theta\,\Vert\,\rho_{\theta_0}\right) \\
&= \mathbb{E}_{q\sim\rho_\theta}\!\Big[\log\tfrac{\rho_\theta(q)}{\rho_{\theta_0}(q)}\Big].
\end{split}
\end{equation}
Empirically (Figure~\ref{fig:kl-loss-comparison}), under a fixed action-level KL budget the batch-estimated Query-KL keeps rising throughout training while the response-level Policy-KL stays nearly flat---constraining only the action distribution does not stabilize the input/query process. This motivates the two components in Section~\ref{sec:method}: a query-level KL that bounds environment drift, and a per-query weighting scheme that stabilizes the empirical objective under arbitrary pipeline biases.

%% file: tex/method_2.tex
\section{Method}
\label{sec:method}

We treat the pre-RL model-induced query distribution $\rho_{\theta_0}$ as the
\emph{reference environment}. As $\theta$ changes, $\rho_\theta$ may drift
while the exogenous $\rho_{\mathrm{train}}$ remains fixed; our
\emph{Environment-Regularized Policy Optimization} (ERPO) bounds the
former while leaving response-side exploration nearly
unconstrained\footnote{See Section~\ref{sec:method-reweight-qkl} for
details. The only residual constraint stems from coupling among the LLM
parameters. Nevertheless, Figure~\ref{fig:dynamic_240} and the
evaluation results suggest that the parameter space of the 7B model
contains approximately orthogonal subspaces, such that constraining
query-KL does not materially interfere with policy convergence.}.

\paragraph{Working assumption (A1).}
We treat as a premise---not a theorem---that maintaining alignment between
the model-induced $\rho_\theta$ and its pre-RL reference $\rho_{\theta_0}$
better preserves the generalization capabilities inherited from prior
training stages than allowing unconstrained drift. A1 is examined empirically
in Section~\ref{sec:experiments}.

\paragraph{Query likelihood.}
Our method relies on the autoregressive sequence likelihood the model assigns to a query. For a query $q$ tokenized as $(x_1,\dots,x_T)$, write $P_\theta(q) \triangleq \prod_t P_\theta(x_t\mid x_{<t})$ for this sequence likelihood and $\ell_\theta(q) \triangleq \log P_\theta(q) = \sum_t \log P_\theta(x_t\mid x_{<t})$ for its log form, with $P_{\theta_0}(q)$ and $\ell_{\theta_0}(q)$ defined analogously under the reference.

The reference $\ell_{\theta_0}$ is computed once over the training set and cached, while $\ell_\theta$ is read off the per-step PG forward pass; the cached $\ell_{\theta_0}$ table and the per-step $\ell_\theta$ are the only inputs ERPO needs beyond what the underlying PG estimator already computes (see Appendix~\ref{app:likelihood-details} for full computational notes).

\subsection{Population objective and empirical loss}
\label{sec:method-obj}

Let $\bar g_\theta(q):=\mathbb{E}_{o\sim\pi_\theta(\cdot\mid q)}[g(q,o)]$
be the per-query expected reward, so that
$J_{\mathrm{RL}}(\theta)=\mathbb{E}_{q\sim\rho_{\mathrm{train}}}[\bar g_\theta(q)]$.
ERPO adds a query-level regularizer $\mathcal{R}_{\mathrm{query}}(\theta)$
penalizing the divergence between the model-induced $\rho_\theta$ and
$\rho_{\theta_0}$:
\begin{equation}
\label{eq:method-J-erpo}
J_{\mathrm{ERPO}}(\theta)
:=
J_{\mathrm{RL}}(\theta)
- \alpha\,\mathcal{R}_{\mathrm{query}}(\theta),
\end{equation}
where $\alpha>0$ controls regularization strength. On a mini-batch $B=\{q_i\}_{i=1}^{m}$, we additionally reweight per-query contributions by a dataset-static, reference-derived weight $w_B(q)$ (Section~\ref{sec:method-reweight-qkl}); together with a batch-level KL estimate $\widehat{\mathcal{R}}_{\mathrm{query}}(\theta)$, this gives the empirical loss
\begin{equation}
\label{eq:method-Japprox}
\begin{split}
\widehat{L}_{\mathrm{ERPO}}(\theta)
:=
&-\frac{1}{m}\sum_{q\in B}w_B(q)\,\bar g_\theta(q) \\
&+\alpha\,\widehat{\mathcal{R}}_{\mathrm{query}}(\theta).
\end{split}
\end{equation}
$w_B$ enters only at the estimator level; it shapes the SGD update direction toward queries typical under $\rho_{\theta_0}$ but does not enter the target $J_{\mathrm{ERPO}}$.

\subsection{Query-level KL and query reweighting}
\label{sec:method-reweight-qkl}

\paragraph{Query-level KL.}
We instantiate the regularizer as the KL from the current query distribution to the reference:
\begin{equation}
\label{eq:method-qkl}
\begin{split}
\mathcal{R}_{\mathrm{query}}(\theta)
&:= \mathrm{KL}\!\left(\rho_\theta\,\big\Vert\,\rho_{\theta_0}\right) \\
&= \mathbb{E}_{q\sim\rho_\theta}\big[\log\rho_\theta(q)-\log\rho_{\theta_0}(q)\big].
\end{split}
\end{equation}
QKL \emph{decouples regularization from exploration}: the penalty acts solely on the query distribution through $\ell_\theta(q)$, while imposing no direct constraint on $\pi_\theta(o\mid q)$. Conventional Policy-KL, in contrast, restricts response behavior and consumes the action-side exploration budget.

\textbf{Proposition 1 (Structural Decoupling).}
\emph{For autoregressive $\pi_\theta$, the gradient of $\mathcal{R}_{\mathrm{query}}(\theta)$ admits the closed form}
\begin{equation}
\label{eq:method-qkl-grad}
\begin{split}
\nabla_\theta\mathcal{R}_{\mathrm{query}}(\theta)
=\mathbb{E}_{q\sim\rho_\theta}\!\Big[
&\bigl(\ell_\theta(q)-\ell_{\theta_0}(q)\bigr) \\
&\cdot\nabla_\theta\ell_\theta(q)\Big],
\end{split}
\end{equation}
\emph{which flows strictly through $\nabla_\theta\ell_\theta(q)$; the response score function $\nabla_\theta\log\pi_\theta(o\mid q)$ used by PG estimators does not appear in the QKL loss, so the regularizer exerts no direct gradient pressure on the response distribution. Proof in Appendix~\ref{app:proof-decoupling}.}
In our fixed-dataset implementation, we evaluate a K3-style mini-batch
surrogate using the per-step $\ell_\theta(q)$ and cached
$\ell_{\theta_0}(q)$. It regularizes model-induced query-likelihood drift
without treating model likelihood as external query frequency. The resulting
$\widehat{\mathcal{R}}_{\mathrm{query}}(\theta)$ is used in
Eq.~\eqref{eq:method-Japprox}.

\input{tex/wq_short}


\paragraph{PG-compatible surrogate.}
ERPO is agnostic to the inner PG estimator: any surrogate of the form $\nabla_\theta\bar g_\theta(q)\approx\mathbb{E}_o[u_\theta(q,o)A_\theta^\star(q,o)\nabla_\theta\log\pi_\theta(o\mid q)]$ recovers GRPO ($u_\theta\equiv 1$, $A^\star$ from Eq.~\eqref{eq:prelim-grpo-adv}), PPO (clipped ratios), or REINFORCE (sample reward) by appropriate choice of $u_\theta,A_\theta^\star$. Plugging into Eq.~\eqref{eq:method-Japprox} gives the ERPO mini-batch PG loss
\begin{equation}
\label{eq:method-pg-surrogate}
\begin{split}
\mathcal{L}_{\mathrm{PG}}(\theta)
:=
&-\frac{1}{m}\sum_{q\in B}\frac{w_B(q)}{K}\sum_{o\in\mathcal{G}(q)}u_\theta(q,o)\,A_\theta^\star(q,o) \\
&+\alpha\,\widehat{\mathcal{R}}_{\mathrm{query}}(\theta),
\end{split}
\end{equation}
where $\mathcal{G}(q)$ is the $K$-response group sampled for $q$. Per-algorithm specializations (GRPO/PPO/REINFORCE) and the resulting loss decompositions are summarized in Appendix~\ref{app:pg-details}.


%% file: tex/wq_short.tex
\paragraph{Query reweighting.}
Instead of optimizing the standard RL objective in Eq.~\eqref{eq:rl-objective} under the empirical query distribution
\(\rho_{\mathrm{train}}\), ERPO targets the reference-aligned query prior
\(\rho_{\theta_0}\):
\begin{equation}
\label{eq:erpo-objective-rhotheta0}
J_{\mathrm{ERPO}}(\theta)
=
\mathbb{E}_{q\sim{\color{red}\rho_{\theta_0}},\,o\sim\pi_\theta(\cdot\mid q)}
[g(q,o)].
\end{equation}
Because stochastic training samples queries from
\(\rho_{\mathrm{train}}\), this objective can be written as an importance-weighted objective with ideal weight
\begin{equation}
\label{eq:ideal-query-weight}
w^\star(q)=\frac{\rho_{\theta_0}(q)}{\rho_{\mathrm{train}}(q)}.
\end{equation}
Under approximately uniform sampling from a dataset of size \(N\),
\(w^\star(q)=N\rho_{\theta_0}(q)\), so the query weight only needs to preserve the relative ordering of the cached reference-induced query prior. We therefore use a bounded, dataset-static weight
\(w(q)\propto\ell_{\theta_0}(q)\), leading to
\begin{align}
\label{eq:erpo-loss-query-weighted}
L_{\mathrm{ERPO}}(\theta)
&=
-
\mathbb{E}_{\substack{
q\sim\rho_{\mathrm{train}}\\
o\sim\pi_\theta(\cdot\mid q)
}}
\left[
w(q)g(q,o)
\right]
\nonumber \\
&\quad
+
\alpha\mathcal{R}_{\mathrm{query}}(\theta).
\end{align}
The derivation is given in Appendix~\ref{app:query-reweighting-derivation}.

%% file: tex/experiments.tex
\section{Experiments}
\label{sec:experiments}

\input{tex/main_result_figure}

\input{tex/main_result_table}

\subsection{Experimental Setup}
\paragraph{Training} We conduct experiments on mathematical reasoning tasks using Level~3--5 problems from the MATH dataset~\citep{Hendrycks_Burns_Kadavath_Arora_Basart_Tang_Song_Steinhardt_2021}, totaling approximately 8.5K examples. These are used to evaluate our proposed ERPO method, in comparison with the vanilla GRPO baseline. As described in Appendix~\ref{sec:app_prompt}, the model must wrap its intermediate reasoning in \verb|<think></think>| tags, and place the final answer inside \verb|\boxed {}|.

\paragraph{Evaluation} We follow standard practice and assess performance on six widely used benchmarks: AIME24, AIME25, AMC, MATH500~\citep{Hendrycks_Burns_Kadavath_Arora_Basart_Tang_Song_Steinhardt_2021}, Minerva~\citep{lewkowycz2022solving}, and OlympiadBench~\citep{he2024olympiadbench}. Prior work typically reports Avg@K~\citep{yu2025dapo}, Pass@1~\citep{liu2025understanding}, and Pass@K~\citep{hao2025policy} after RLVR training, often without specifying or controlling the inference-time sampling temperature. This omission can substantially affect reported performance and render results across studies not directly comparable. In preliminary experiments, we found that inference-time sampling temperature has a significant impact on performance, and that the effect intensifies as training progresses. To control for this factor, we fix the number of training steps across all models and evaluate at temperatures from 0.1 to 1.5; performance is then aggregated over this range.

\paragraph{Implementation Details}
We conduct all experiments using the EasyR1 framework \citep{zheng2025easyr1}, training the Qwen2.5-Math-7B and Qwen2.5-32B model \citep{yang2024qwen2, qwen2025qwen25technicalreport} with both GRPO and ERPO algorithms. Following prior work \citep{liu2025understanding}, we set the maximum sequence length to 3K tokens. For each problem, we sample eight responses at an inference temperature of 1.0. The rollout batch size is set to 512, and the update batch size to 128, for a total of 240 training steps. Token-level loss is applied throughout training. To ensure a fair comparison, we adopt the default KL divergence coefficient of 0.01.

\subsection{Main Results}
Figure~\ref{fig:mainresult} summarizes Avg@32 accuracy on six mathematical reasoning benchmarks, averaged over sampling temperatures from 0.1 to 1.5. ERPO consistently outperforms GRPO, with gains of up to 14.9\% and an overall average improvement of 6.2\%, highlighting its enhanced capability. Table~\ref{tab:main} presents the detailed results for each benchmark, grouped by evaluation metric (e.g., Pass@1, Pass@K).

For both GRPO and ERPO, the prompts are identical to those used during training, whereas the Qwen base model adopts the default configuration from Dr.GRPO~\citep{liu2025understanding} to ensure optimal performance. Consistent with the aggregated results in Figure~\ref{fig:mainresult} and Table~\ref{tab:main}, ERPO surpasses GRPO across all evaluation metrics, achieving improvements of 6.2\% in Avg@32, 3.64\% in Pass@32, and 5.69\% in Pass@1. We also applied the concept of ERPO to other RLVR algorithms and observed similarly effective gains; details are provided in Appendix~\ref{sec:algorithms}.

\begin{figure*}[!t]
  \centering
  \resizebox{0.95\textwidth}{!}{    \includegraphics{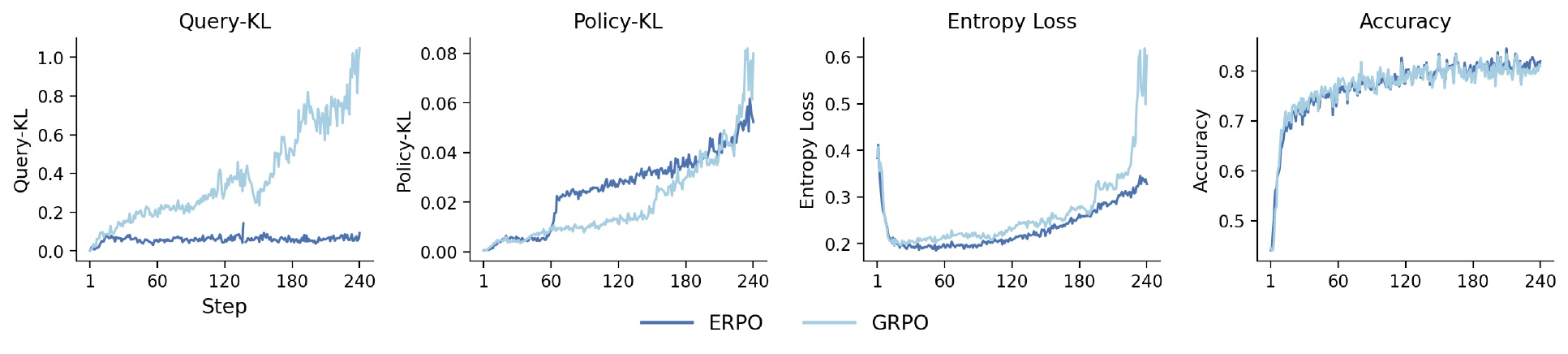}
  }
  \vspace{-5pt}
  \caption{Training Dynamics on ERPO}
  \label{fig:dynamic_240}
  \vspace{1pt}
\end{figure*} 

\begin{figure*}[!t]
  \centering
  \resizebox{0.95\textwidth}{!}{
    \includegraphics{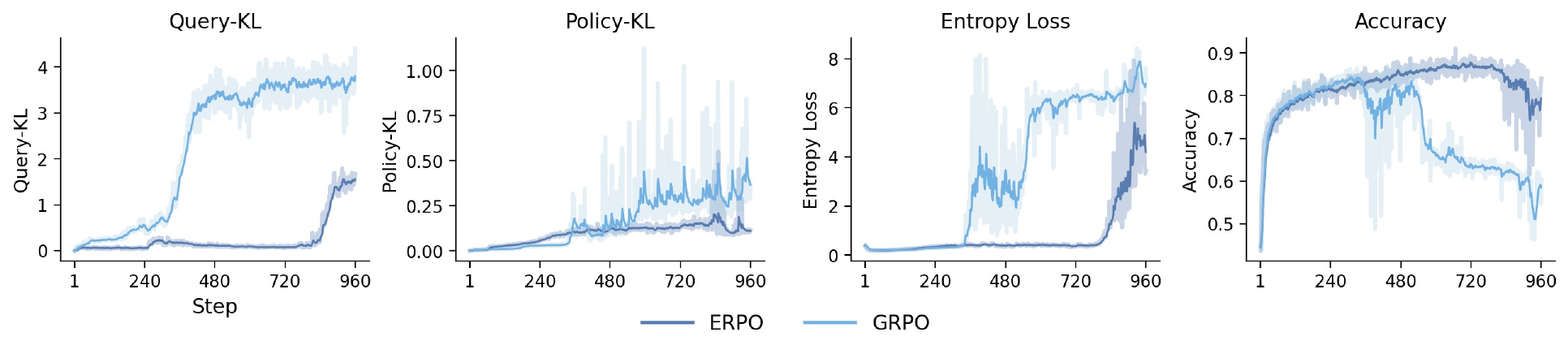}
  }
  \vspace{-5pt}
  \caption{Training Dynamics on Long-term RL}
  \label{fig:dynamic_960}
  \vspace{1pt}
\end{figure*}

\subsection{Training Dynamics}

Figure \ref{fig:dynamic_240} illustrates the training dynamics of the ERPO method. For both approaches, the sampling accuracy on the training set remains largely consistent; however, their divergence from the reference model exhibits markedly different trajectories. 

In GRPO, constraints are imposed on the action distribution, causing the query distribution to drift away from the reference model at a substantially faster rate. Consequently, the KL divergence at the query level is an order of magnitude greater than at the policy level.

This imbalance leads to pronounced discrepancies in performance between the training and evaluation datasets. In contrast, ERPO applies constraints directly to the query distribution and adjusts the loss according to the probability of the given problem. This design both limits the degree of divergence from the reference model during training and, by leveraging the independence between the problem and the response, allows unconstrained exploration at the policy level. A quantitative analysis can be found in Appendix~\ref{sec:hacking}. As a result, ERPO achieves superior generalization performance on general problems.

\begin{figure}[!t]
  \centering
  \resizebox{0.47\textwidth}{!}{
    \includegraphics{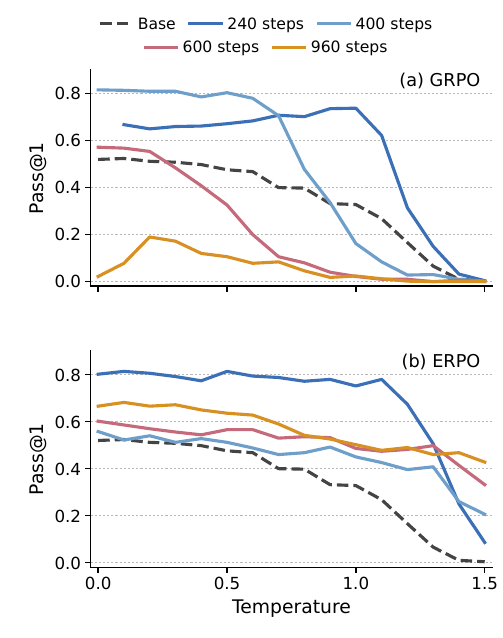}
  }
  \vspace{-5pt}
  \caption{Performance Variation Across Training Steps}
  \label{fig:longterm}
  \vspace{1pt}
\end{figure}

To assess the stability of long-term RL training, we scale the training steps up to 1K and monitor changes in model performance over time. As shown in the figure \ref{fig:dynamic_960} and \ref{fig:longterm}, GRPO remains stable for sampling temperatures below 1.0 until approximately 240 steps (epoch=15). However, a pronounced performance degradation is first observed in the high-temperature sampling regime after 400 steps, and subsequently propagates to encompass sampling across all temperatures as the steps increase.

In contrast, ERPO exhibits a modest performance decline; however, the overall deterioration is substantially smaller, and its performance even improves within the high-temperature range. Figure \ref{fig:longterm} presents the complete training trajectories for both GRPO and ERPO. Although ERPO is not entirely immune to the collapse phenomenon that may occur during extended training—manifested as a sudden increase in entropy and a loss of sampling capability—it consistently outperforms vanilla GRPO and achieves a comparable degree of policy distribution constraint without relying on an explicit policy-based KL divergence term.

\input{tex/table_xiaorong}

\begin{table}[ht]
\centering
\caption{Influence of Query-KL and Query-Reweighting on Training Stability}
\label{tab:kl_entropy}
\footnotesize  
\begin{tabular}{cccc}
\toprule
\textbf{Method} & \textbf{Query-KL} & \textbf{Policy-KL} & \textbf{Entropy} \\
\midrule
GRPO (Avg@3)            & 0.9679  & 0.0601           & 0.5063 \\
GRPO w/$w_{B(s)}$     & 0.5933  & \textbf{0.0113}  & \textbf{0.2782} \\
GRPO w/Query-KL   & \textbf{0.0041} & 0.1001 & 0.5674 \\
ERPO (Avg@3)            & 0.0828  & 0.0728           & 0.4244 \\
\bottomrule
\end{tabular}
\end{table}

\begin{figure*}[!t]
  \centering
  \resizebox{0.98\textwidth}{!}{
    \includegraphics{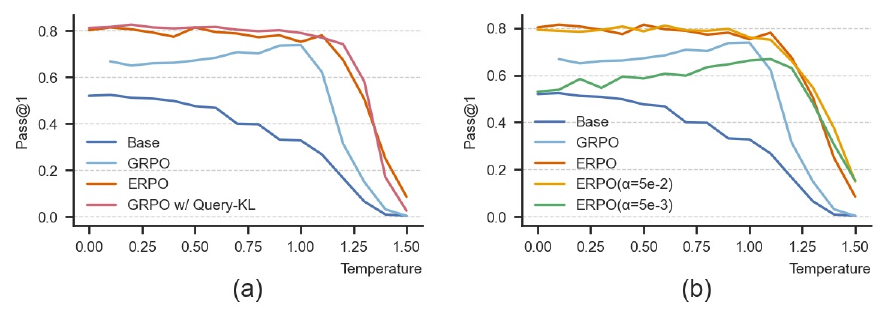}
    \includegraphics{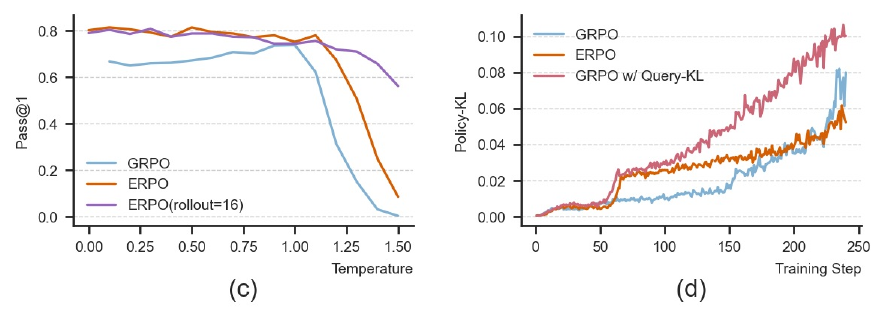}
  }
  \vspace{-5pt}
  \caption{Pass@1 accuracy and training dynamics under different settings: (a)–(c) Model performance at various temperatures on MATH500; (d) Policy‑KL divergence variation with GRPO using only Query‑KL.}
  \label{fig:ablation}
  \vspace{1pt}
\end{figure*}

\subsection{Analysis}
\paragraph{Ablation Study}
We conduct ablation studies on the MATH500 benchmarks to assess reasoning efficiency. Table \ref{tab:method_comparison} summarizes the results for several commonly used sampling temperatures. Figure \ref{fig:ablation} further provides the complete performance–temperature variation curves across different experimental settings, along with the corresponding training dynamics.

\paragraph{Mechanisms}
Overall, Query-KL is the primary driver of the performance gains,
whereas query weighting (QW) mainly plays a complementary role in
stabilizing training. Specifically, without modifying other
hyperparameters, replacing the policy-based KL divergence with
query-based KL divergence yields the best overall
performance\footnote{We also experimented with completely removing all
KL divergence constraints, which resulted in the training process
failing to converge. Therefore, we only report GRPO results under the
default-weight KL divergence constraint.}, with an average improvement
of 15.9\% over GRPO. In contrast, GRPO with policy-based KL divergence
achieves its highest performance only at a temperature of 1.0 (see
Figure~\ref{fig:ablation}(a)).

To further investigate, an ablation study is conducted on the two
mechanisms of ERPO, with their effects evaluated using KL divergence
and entropy (Table~\ref{tab:kl_entropy} and
Figure~\ref{fig:ablation}(d)). The term $w_{B(s)}$ downweights gradients
from low-probability queries, which often yield low-probability
responses and consequently increase gradient variance and entropy
(a quantitative analysis is provided in
Appendix~\ref{sec:corr}). Introducing $w_{B(s)}$ enables sufficient
training while concurrently reducing the policy KL divergence. In
contrast, replacing the original policy-KL term with query-KL alone
increases policy entropy, thereby enhancing the model's capacity for
exploration.

Different regularization strengths $\alpha$ also exert a significant influence on performance. As the constraint strength increases (e.g., $\alpha=5 \times 10^{-2}$), the model achieves further improvements in overall performance (see Table \ref{tab:method_comparison}). It is worth noting that we did not conduct an exhaustive search for the optimal $\alpha$; instead, we retained the default value to ensure a relatively fair comparison.

\paragraph{Rollouts}
We also analyze the effect of the number of samples per query.
By increasing the sampling number to 16, we achieve the best performance, 
with the average Pass@1 rising to $74.6\%$. 
A higher sampling count also significantly improves sampling stability at high temperatures (see Table~\ref{tab:method_comparison}), 
without a noticeable increase in divergence from the reference model. Moreover, increasing the sampling count facilitates ERPO-based models in acquiring the correct reasoning format more effectively.
\footnote{Across multiple experiments, the GRPO method consistently failed to capture the desired output format. Consequently, for all experiments, we report only the answer accuracy.}

%% file: tex/main_result_figure.tex
\begin{figure*}[!t]
  \centering
  \resizebox{0.8\textwidth}{!}{\includegraphics{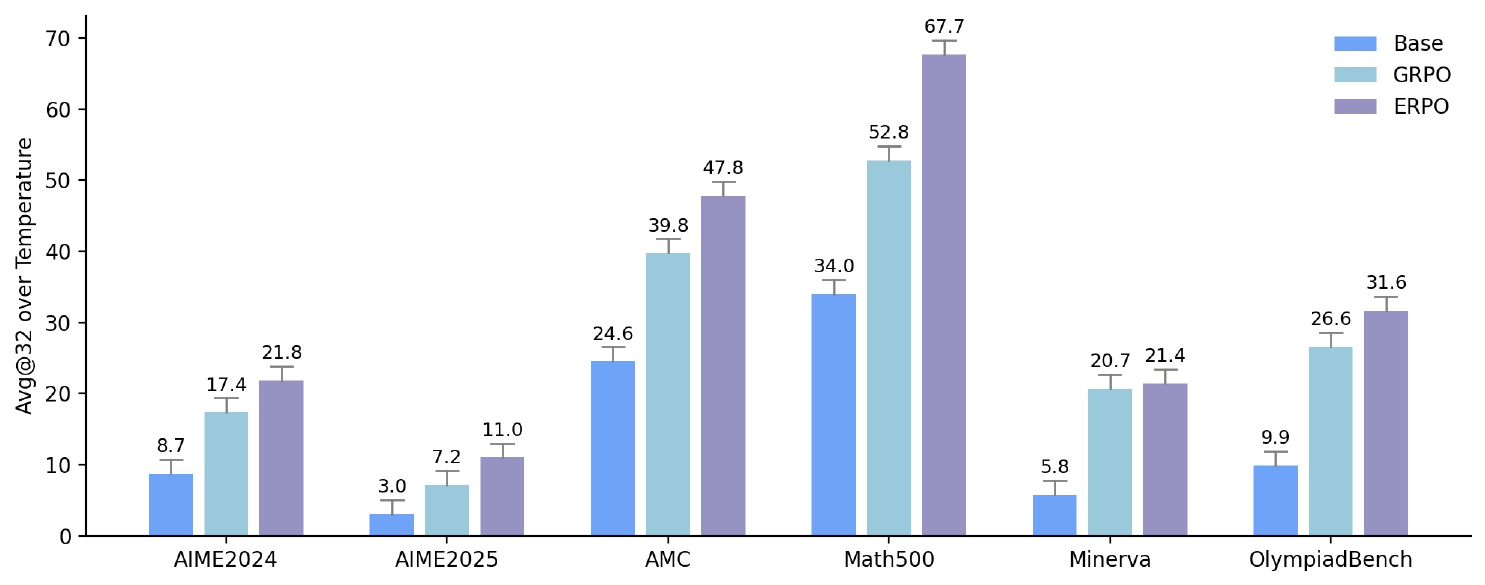}}
  \vspace{-5pt}
  \caption{Avg@32 over Sampling Temperatures on Mathematical Reasoning Tasks}
  \label{fig:mainresult}
  \vspace{1pt}
\end{figure*}

%% file: tex/main_result_table.tex
\begin{table*}[htbp]
\centering
\setlength{\belowcaptionskip}{10pt}  
\caption{Performance comparison across mathematical reasoning benchmarks. Best results per column are highlighted in bold.}
\label{tab:main}
\renewcommand{\arraystretch}{1.2}
\setlength{\tabcolsep}{4pt}
\footnotesize  
\begin{tabular}{l|ccccccc}
\toprule
\multicolumn{8}{c}{\textbf{Mean Avg@32}} \\
\midrule
Method & \textit{AIME24} & \textit{AIME25} & \textit{AMC} & \textit{MATH500} & \textit{Minerva} & \textit{Olympiad} & Avg. \\
\midrule
Base & 0.087 & 0.030 & 0.246 & 0.340 & 0.058 & 0.099 & 0.143 \\
GRPO & 0.174 & 0.072 & 0.398 & 0.528 & 0.207 & 0.266 & 0.274 \\
ERPO & \textbf{0.218} & \textbf{0.110} & \textbf{0.478} & \textbf{0.677} & \textbf{0.214} & \textbf{0.316} & \textbf{0.336} \\
\midrule
\multicolumn{8}{c}{\textbf{Mean Pass@32}} \\
\midrule
Base & 0.373 & 0.206 & 0.674 & 0.764 & 0.349 & 0.411 & 0.463 \\
GRPO & 0.471 & 0.287 & 0.768 & 0.850 & \textbf{0.516} & 0.558 & 0.575 \\
ERPO & \textbf{0.509} & \textbf{0.342} & \textbf{0.820} & \textbf{0.904} & 0.500 & \textbf{0.593} & \textbf{0.611} \\
\midrule
\multicolumn{8}{c}{\textbf{Mean Pass@1}} \\
\midrule
Base & 0.090 & 0.038 & 0.264 & 0.342 & 0.062 & 0.099 & 0.149 \\
GRPO & 0.169 & 0.084 & 0.398 & 0.533 & 0.201 & 0.263 & 0.275 \\
ERPO & \textbf{0.207} & \textbf{0.091} & \textbf{0.477} & \textbf{0.679} & \textbf{0.217} & \textbf{0.320} & \textbf{0.332} \\
\bottomrule
\end{tabular}
\end{table*}

%% file: tex/table_xiaorong.tex
\begin{table*}[htbp]
\centering
\caption{Performance Comparison Under Different Experimental Settings}
\label{tab:method_comparison}
\small
\setlength{\tabcolsep}{2.5pt}

\begin{tabular}{@{}lp{2.8cm}ccccccccc@{}}
\toprule
\multirow{2}{*}{\textbf{Base Model}}
& \multirow{2}{*}{\textbf{Method}}
& \multirow{2}{*}{$\boldsymbol{\alpha}$}
& \multirow{2}{*}{\textbf{$w(s)$}}
& \multirow{2}{*}{\shortstack{\textbf{Rollout}\\\textbf{Count}}}
& \multicolumn{6}{c}{\textbf{Temperature Metrics}} \\
\cmidrule(lr){6-11}
& & & &
& \textbf{0.1}
& \textbf{0.6}
& \textbf{1}
& \textbf{1.5}
& \textbf{$\leq 1.0$}
& \textbf{1.2--1.5} \\
\midrule

\textbf{Baseline}
& ---
& ---
& ---
& ---
& 52.40
& 46.80
& 32.80
& 0.40
& 44.44
& 6.15 \\

\midrule

\multirow{6}{*}{\textbf{Qwen-7B}}
& GRPO
& $1 \times 10^{-2}$
& ---
& 8
& 66.80
& 68.40
& 73.80
& 0.40
& 68.80
& 12.50 \\

& GRPO w/ $w(s)$
& $1 \times 10^{-2}$
& \checkmark
& 8
& 78.20
& 76.80
& 71.20
& 7.60
& 76.14
& 23.79 \\

& GRPO w/ query-KL
& $1 \times 10^{-2}$
& ---
& 8
& \textbf{81.60}
& \textbf{81.60}
& \textbf{79.00}
& 2.60
& \textbf{80.90}
& 38.00 \\

\cmidrule(lr){2-11}

& \multirow{3}{*}{ERPO}
& $1 \times 10^{-2}$
& \checkmark
& 8
& \underline{79.40}
& 80.60
& 75.20
& 8.60
& 78.74
& 37.90 \\

& &
$5 \times 10^{-3}$
& \checkmark
& 8
& 53.80
& 60.60
& 66.20
& \textbf{15.40}
& 59.94
& \underline{39.30} \\

& &
$5 \times 10^{-2}$
& \checkmark
& 8
& 78.80
& \underline{81.00}
& \underline{76.00}
& \underline{15.00}
& \underline{79.00}
& \textbf{43.35} \\

\midrule

\multirow{2}{*}{\shortstack[l]{\textbf{Qwen-7B}\\\textbf{under $n=16$}}}
& GRPO
& $1 \times 10^{-2}$
& ---
& 16
& 73.00
& 79.20
& 75.00
& 10.60
& 75.22
& 39.75 \\

& ERPO
& $1 \times 10^{-2}$
& \checkmark
& 16
& 80.40
& 78.80
& 74.40
& 56.20
& 77.82
& 66.25 \\

\midrule

\multirow{2}{*}{\textbf{Qwen-32B}}
& GRPO
& $1 \times 10^{-2}$
& ---
& 8
& 81.60
& 82.40
& 81.20
& 25.20
& 81.62
& 57.20 \\

& ERPO
& $1 \times 10^{-2}$
& \checkmark
& 8
& 85.00
& 84.80
& 83.60
& 80.80
& 84.60
& 82.80 \\

\bottomrule
\end{tabular}

\vspace{5pt}

\begin{minipage}{0.9\textwidth}
\footnotesize
\textbf{Note:}
Within the Qwen-7B ($n=8$) setting, the best result in each column is
highlighted in \textbf{bold}, while the second-best result is
\underline{underlined}. No highlighting is applied to the Qwen-7B
($n=16$) or Qwen-32B results. The first three Qwen-7B rows present an
ablation study comparing standard GRPO, GRPO augmented with query
reweighting $w(s)$, and a GRPO variant in which the original policy-KL
term is replaced by query-KL (denoted as GRPO w/ query-KL). The
following three rows constitute the hyperparameter study, evaluating
ERPO with query reweighting under different values of $\alpha$. The
columns \textbf{$\leq 1.0$} and \textbf{1.2--1.5} report the mean
accuracy (Acc) over the corresponding temperature ranges. The $w(s)$
column indicates whether query reweighting is applied (\checkmark) or
not (---).
\end{minipage}
\end{table*}

%% file: tex/appendix.tex
\section{Prompt}
\label{sec:app_prompt}
{\ttfamily\itshape
\detokenize{{{ content | trim }} You FIRST think about the reasoning process as an internal monologue and then provide the final answer. The reasoning process MUST BE enclosed within <think> </think> tags. The final answer MUST BE put in \boxed{}.}
}

\section{Variation of Metrics with Temperature}
\label{sec:metrics}

\begin{figure*}[!t]
  \centering
  \resizebox{0.95\textwidth}{!}{
    \includegraphics{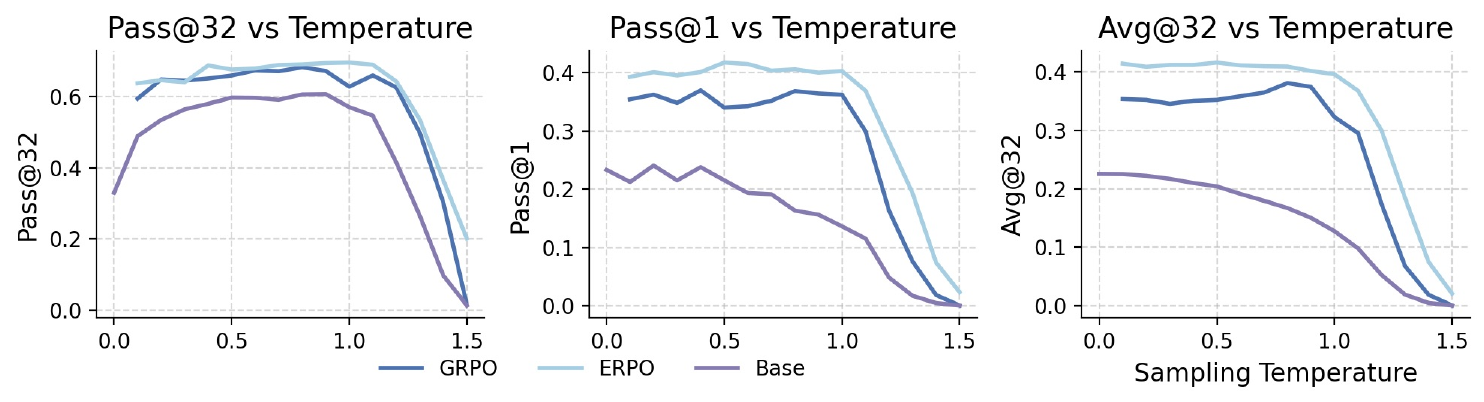}
  }
  \vspace{-5pt}
  \caption{Variation of Metrics with Temperature}
  \label{fig:metrics}
  \vspace{1pt}
\end{figure*}

Figure \ref{fig:metrics} illustrates the model performance across different evaluation metrics and sampling temperatures. Our approach reduces the performance gap between different sampling temperatures, while increasing the likelihood of sampling correct outputs.

\section{Analysis of Reward Hacking}
\label{sec:hacking}
In the course of our experiments, we observed a severe reward hacking phenomenon when training with the baseline GRPO method. Specifically, while the model consistently achieved high rewards on the training data, its performance on the evaluation set often plateaued or even degraded during the later stages of optimization. This pronounced discrepancy suggests that the model overfits to the specific characteristics of the reward signal during training sampling, failing to generalize to the standard decoding distribution used during inference.

To quantitatively investigate this issue, we monitored the Train–Evaluation Consistency throughout the training process. We periodically evaluated both the training accuracy and the evaluation accuracy every ten optimization steps. To ensure the robustness of our inference metrics, evaluation was conducted using vLLM under two different Tensor Parallelism settings (TP1 and TP2), a factor which has been shown in previous work \citep{yao2025offpolicy} to impact model performance.

\begin{table*}[ht]
    \centering
    \caption{\textbf{Quantification of Reward Hacking via Train–Inference Gap.} The table compares the average accuracy during training sampling versus inference decoding. A larger gap indicates severe reward hacking (overfitting to training dynamics). ERPO reduces this gap by $\approx 51\%$, demonstrating robust generalization.}
    \label{tab:avg_gap_summary}
    \resizebox{\linewidth}{!}{%
    \begin{tabular}{lcccccc}
        \toprule
        \textbf{Method} & \textbf{Avg. Train Acc} & \textbf{Avg. Eval@TP1} & \textbf{Avg. Eval@TP2} & \textbf{Gap@TP1} & \textbf{Gap@TP2} & \textbf{Avg. Gap} \\
        \midrule
        GRPO & 77.3 & 69.8 & 70.1 & 7.5 & 5.5 & 6.47 \\
        ERPO & 77.1 & 74.1 & 73.8 & 3.0 & 3.3 & 3.14 \\
        \midrule
        \textbf{Improvement} & -- & -- & -- & \textcolor{blue}{$\downarrow 4.5$ (60\%)} & \textcolor{blue}{$\downarrow 2.2$ (40\%)} & \textcolor{blue}{$\downarrow 3.33$ (51\%)} \\
        \bottomrule
    \end{tabular}
    }
\end{table*}

Table \ref{tab:avg_gap_summary} summarizes the average performance gap across six key checkpoints (Steps 40, 80, 120, 160, 200, and 240). The results confirm our hypothesis:
\begin{itemize}
\item GRPO exhibits a substantial average gap of $6.47\%$, indicating a significant misalignment between training and inference behaviors. Notably, as shown in the detailed trajectories in Table \ref{tab:detailed_step_acc}, GRPO's evaluation accuracy drops sharply at Step-240 (from $\approx 75\%$ to $58.4\%$) despite maintaining high training accuracy, a classic signature of reward hacking.
\item ERPO, in contrast, demonstrates superior consistency. It reduces the average Train–Eval gap by approximately 51\% (from $6.47\%$ to $3.14\%$).
\end{itemize}

\begin{table*}[ht]
    \centering
    \caption{\textbf{Trajectory of Train vs. Inference Accuracy.} Detailed performance recorded at 40-step intervals. Note the divergence in GRPO at Step-240, where Eval accuracy drops significantly while Train accuracy remains high—a clear sign of reward hacking. ERPO maintains consistency throughout.}
    \label{tab:detailed_step_acc}
    \resizebox{\linewidth}{!}{%
    \begin{tabular}{lcccccccc}
        \toprule
        \textbf{Model / Metric} & \textbf{Step-0} & \textbf{Step-40} & \textbf{Step-80} & \textbf{Step-120} & \textbf{Step-160} & \textbf{Step-200} & \textbf{Step-240} & \textbf{Avg.} \\
        \midrule
        \multicolumn{9}{l}{\textit{GRPO (Baseline)}} \\
        \quad Train Acc & 44.48 & 76.41 & 76.09 & 78.29 & 78.95 & 79.44 & 76.70 & 72.91 \\
        \quad Eval@TP1 & 31.20 & 73.20 & 76.00 & 73.40 & 72.20 & 73.60 & 58.40 & 65.43 \\
        \quad Eval@TP2 & 30.60 & 74.00 & 74.80 & 74.20 & 76.60 & 75.80 & 66.20 & 67.46 \\
        \midrule
        \multicolumn{9}{l}{\textit{ERPO (Ours)}} \\
        \quad Train Acc & 44.55 & 75.63 & 76.53 & 78.66 & 80.63 & 78.41 & 81.46 & 73.70 \\
        \quad Eval@TP1 & 31.20 & 74.00 & 77.20 & 77.00 & 78.40 & 78.60 & 78.40 & 70.69 \\
        \quad Eval@TP2 & 30.60 & 72.60 & 77.00 & 76.60 & 77.40 & 78.60 & 80.20 & 70.43 \\
        \bottomrule
    \end{tabular}
    }
\end{table*}

This significant reduction in the performance gap indicates that ERPO effectively regularizes the training process, preventing the model from exploiting spurious patterns in the reward function and ensuring that improvements in training translate reliably to inference performance.

\section{Correlation analysis between query and response probabilities}
\label{sec:corr}
\begin{figure*}[!t]
  \centering
  \resizebox{0.95\textwidth}{!}{
    \includegraphics{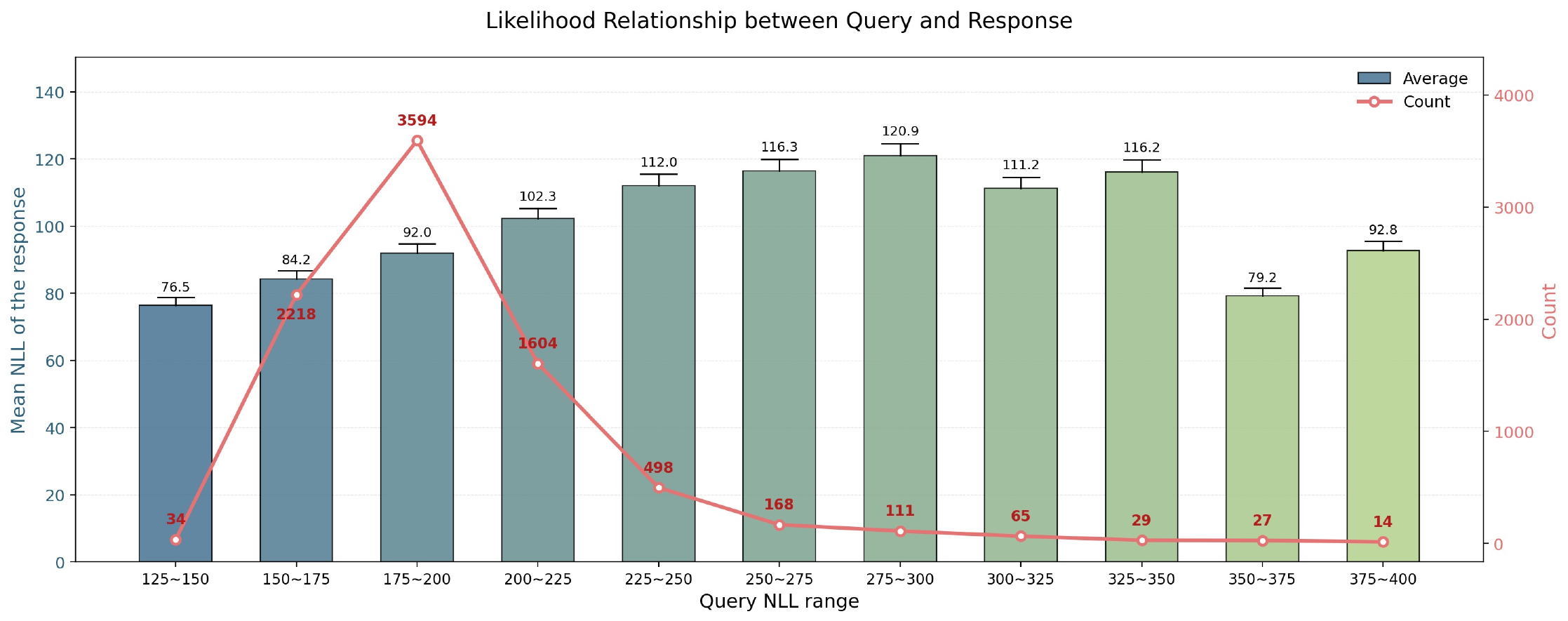}
  }
  \vspace{-5pt}
  \caption{Likelihood Relationship between Query and Response }
  \label{fig:likelihood}
  \vspace{1pt}
\end{figure*}

We sampled over 8K questions from the training dataset at a temperature of 1.0 and independently computed the negative log-likelihood (NLL) for both the prompt and response parts:
$$
\text{NLL}(x) = -\sum \log p(x)
$$
where $x$ denotes the generation probabilities of tokens in the prompt or response. The resulting histogram is shown in Figure~\ref{fig:likelihood}. We observe a positive correlation between the NLL of the prompt and that of the response. For 95\% of the training samples (where the NLL of the prompt is less than 300), the correlation coefficient is close to 1. Low-probability responses appearing in positive samples contribute to an increase in entropy during training.

\section{ERPO On Different Algorithms}
\label{sec:algorithms}
Additional experiments were conducted on DAPO\citep{yu2025dapo} and RLOO\cite{ahmadian2024basicsrevisitingreinforcestyle} with and without the global KL divergence constraint, yielding absolute improvements of 10.24\% and 2.28\% at temperatures below 1.0, respectively (see Table~\ref{tab:algo_comparison}). These findings demonstrate that the proposed method can achieve significant gains when applied to other RLVR algorithms.

\begin{table*}[htbp]
\centering
\caption{Performance Comparison Under Different Experimental Settings}
\label{tab:algo_comparison}
\small
\setlength{\tabcolsep}{4pt}
\begin{tabular}{llcccccccc}
\toprule
\multirow{2}{*}{\textbf{Base Model}} & \multirow{2}{*}{\textbf{Method}} & \multirow{2}{*}{\textbf{D\_KL}} & \multirow{2}{*}{\textbf{Rollout}} & \multicolumn{6}{c}{\textbf{Temperature}} \\
\cmidrule(lr){5-10}
& & & & \textbf{0.1} & \textbf{0.6} & \textbf{1} & \textbf{1.5} & \textbf{$\leq$1.0} & \textbf{1.2-1.5} \\
\midrule
Baseline & -- & -- & -- & 52.4 & 46.8 & 32.8 & 0.4 & 44.44 & 6.15 \\
\midrule
\multirow{9}{*}{Qwen-7B} 
& \multirow{2}{*}{GRPO} & \multirow{2}{*}{Policy} & 8 & 66.8 & 68.4 & 73.8 & 0.4 & 68.8 & 12.5 \\
& & & 16 & 73.0 & 79.2 & 75.0 & 10.6 & 75.22 & 39.75 \\
\cmidrule(lr){2-10}
& \multirow{2}{*}{ERPO} & \multirow{2}{*}{Query} & 8 & 79.4 & {80.6} & 75.2 & 8.6 & 78.74 (+9.94) & 37.9 \\
& & & 16 & {80.4} & 78.8 & 74.4 & \underline{56.2} & {77.82} (+2.6) & \underline{66.25} \\
\cmidrule(lr){2-10}
& DAPO & -- & \multirow{2}{*}{8} & 62.0 & 77.4 & 65.4 & 5.6 & 68.16 & 14.0 \\
& DAPO+ERPO & Query & & {80.2} & 79.4 & 75.8 & 20.2 & 78.4 (+10.24) & 36.93 \\
\cmidrule(lr){2-10}
& RLOO & Policy & \multirow{2}{*}{8} & 77.6 & 78.4 & 75.4 & 12.4 & 77.28 & 35.93 \\
& RLOO+ERPO & Query & & {81.2} & 80.4 & {79.4} & 17.6 & {79.56} (+2.28) & {40.8} \\
\midrule
\multirow{2}{*}{Qwen-32B} 
& GRPO & Policy & \multirow{2}{*}{8} & \underline{81.6} & \underline{82.4} & \underline{81.2} & {25.2} & \underline{81.62} & 57.2 \\
& ERPO & Query & & \textbf{85.0} & \textbf{84.8} & \textbf{83.6} & \textbf{80.8} & \textbf{84.6} (+2.98) & \textbf{82.8} \\
\bottomrule
\end{tabular}
\vspace{8pt}
\begin{minipage}{0.85\textwidth}
\footnotesize
\textbf{Note:} The columns \textbf{$\leq$1.0} and \textbf{1.2--1.5} show the mean accuracy (Acc) over the corresponding temperature ranges. Values in parentheses indicate improvements over baseline methods. Best results per column are highlighted in \textbf{bold}, second-best results are \underline{underlined}.
\end{minipage}
\end{table*}

\input{tex/appendix_method_details}
\input{tex/appendix_derive}
\input{tex/wq_append}


%% file: tex/appendix_method_details.tex
\section{Query Likelihood and the Cached Reference}
\label{app:likelihood-details}

This appendix expands the query likelihood machinery and computational notes referenced from Section~\ref{sec:method}.

\paragraph{Autoregressive sequence likelihood.}
For a query $q$ with tokenization $x=(x_1,\dots,x_T)$, the model's sequence log-likelihood under parameters $\theta$ is
\begin{equation}
\label{eq:app-likelihood}
\ell_\theta(q) \;\triangleq\; \log P_\theta(x) \;=\; \sum_{t=1}^{T} \log P_\theta(x_t \mid x_{<t}).
\end{equation}
This is well-defined for any token sequence, regardless of whether $q$ is sampled on-policy from $\rho_\theta$, drawn from a fixed dataset, or written by a human; computing $\ell_\theta(q)$ requires a single forward pass of the LLM. The reference $\ell_{\theta_0}(q)$ is defined analogously under $\pi_{\theta_0}$.

\paragraph{Computational notes governing ERPO's zero-overhead design.}
Two distinctions in how $\ell_\theta$ and $\ell_{\theta_0}$ are obtained make ERPO essentially free relative to the underlying PG estimator:
\begin{itemize}
    \item $\ell_{\theta_0}(q)$ is computed once over the entire training set using the reference model $\pi_{\theta_0}$, prior to RL, and cached. The per-query table is reused throughout training and does not change. Under backpropagation it is treated as a constant.
    \item $\ell_\theta(q)$ is computed per training step using the current $\theta$. This forward pass is already performed by the underlying PG estimator (for per-query advantage evaluation), so QKL and QW reuse it at no additional forward cost.
\end{itemize}
The cached $\ell_{\theta_0}$ table and the per-step $\ell_\theta$ are the only inputs ERPO needs beyond what the underlying PG estimator already computes. In particular, the query weight $w_B(q)$ from Eq.~\eqref{eq:ideal-query-weight} is fully precomputed and contributes no gradient or extra forward pass; only the QKL estimator $\widehat{\mathcal{R}}_{\mathrm{query}}(\theta)$ uses the per-step $\ell_\theta(q)$, and even there the value is read from the existing PG forward pass.

\section{PG-Compatible Surrogate: Per-Algorithm Specializations}
\label{app:pg-details}

The ERPO mini-batch loss $\mathcal{L}_{\mathrm{PG}}(\theta)$ in Eq.~\eqref{eq:method-pg-surrogate} is instantiated by a particular choice of action-level weight $u_\theta(q,o)$ and advantage $A_\theta^\star(q,o)$, summarized below. The two ERPO modifications---adding the query-level KL term and reweighting the outer per-query sum by $w_B(q)$---are orthogonal to the choice of $(u_\theta, A_\theta^\star)$ and act only on the outer query loop.

\paragraph{GRPO.}
Set $u_\theta \equiv 1$ and $A_\theta^\star = A_\theta^{\mathrm{GRPO}}$ from Eq.~\eqref{eq:prelim-grpo-adv}. Substituting yields
\begin{equation}
\label{eq:app-erpo-grpo}
\begin{split}
\mathcal{L}_{\mathrm{ERPO\text{-}GRPO}}(\theta)
:=
&-\frac{1}{m}\sum_{q\in B}\frac{w_B(q)}{K}\sum_{o\in\mathcal{G}(q)} \\
&\quad A^{\mathrm{GRPO}}_\theta(q,o)\,\log\pi_\theta(o\mid q) \\
&+\alpha\,\widehat{\mathcal{R}}_{\mathrm{query}}(\theta).
\end{split}
\end{equation}
All GRPO engineering details (reward normalization, group size $K$, sampling temperature, etc.) remain unchanged.

\paragraph{PPO.}
We set 
\[
u_\theta(q, o) = \mathrm{clip}\left( \frac{\pi_\theta(o \mid q)}{\pi_{\theta_{\rm old}}(o \mid q)}, \, 1-\epsilon, \, 1+\epsilon \right)
\]
and \(A_\theta^*\) to the (GAE-based or standard) PPO advantage. 
The inner sum recovers the PPO clipped surrogate, 
with the outer \(w_B(q)\) and the added \(\alpha\,\widehat{\mathcal{R}}_{\rm query}\) providing the ERPO wrapping.

\paragraph{REINFORCE.}
Set $u_\theta \equiv 1$ and $A_\theta^\star(q,o) = g(q,o) - b_\theta(q)$ for an arbitrary baseline $b_\theta(q)$; the inner sum recovers the sample-wise REINFORCE estimator.

\paragraph{Drop-in modification recipe.}
Given an existing PG-style training pipeline using any of the above estimators, applying ERPO requires three steps: (i) precompute $\ell_{\theta_0}$ once over the dataset (Appendix~\ref{app:likelihood-details}); (ii) replace the per-query outer weight $1/m$ with $w_B(q)/m$ from Eq.~\eqref{eq:ideal-query-weight}; (iii) add $\alpha\,\widehat{\mathcal{R}}_{\mathrm{query}}(\theta)$ (K3-estimated from the per-step $\ell_\theta$ and the cached $\ell_{\theta_0}$) to the loss. No additional forward passes, no architectural changes, and no modification to the inner PG estimator's clip/baseline logic are required.

%% file: tex/appendix_derive.tex
\section{Proof of Proposition 1: Structural Decoupling of Regularization and Exploration}
\label{app:proof-decoupling}

This appendix provides the formal proof of Proposition~1 stated in Section~\ref{sec:method-reweight-qkl}.

\paragraph{Proposition 1 (restated).}
\emph{For the autoregressive generation process parameterized by $\theta$, the gradient of the Query-KL regularizer $\mathcal{R}_{\mathrm{query}}(\theta)$ strictly flows through the query log-likelihood $\nabla_\theta \ell_\theta(q)$ and is entirely orthogonal to the response policy score function $\nabla_\theta \log \pi_\theta(o\mid q)$.}

\paragraph{Proof.}
By definition,
\begin{equation*}
\mathcal{R}_{\mathrm{query}}(\theta) = \sum_q \rho_\theta(q)\bigl(\log \rho_\theta(q) - \log \rho_{\theta_0}(q)\bigr).
\end{equation*}
Differentiating with respect to $\theta$ and applying the product rule,
\begin{equation}
\label{eq:app-decoupling-step1}
\begin{split}
\nabla_\theta &\mathcal{R}_{\mathrm{query}}(\theta) \\
&= \sum_q \nabla_\theta \rho_\theta(q)\bigl(\log \rho_\theta(q) \\
&\qquad - \log \rho_{\theta_0}(q)\bigr) \\
&\quad + \sum_q \rho_\theta(q)\,\nabla_\theta\bigl(\log \rho_\theta(q) \\
&\qquad - \log \rho_{\theta_0}(q)\bigr).
\end{split}
\end{equation}

\textbf{Second sum collapses.} Since $\rho_{\theta_0}$ does not depend on $\theta$, we have $\nabla_\theta \log \rho_{\theta_0}(q) = 0$. Applying the identity $\nabla_\theta \log \rho_\theta(q) = \nabla_\theta \rho_\theta(q)/\rho_\theta(q)$ to the remaining term gives
\begin{equation}
\label{eq:app-decoupling-step2}
\begin{split}
\sum_q \rho_\theta(q)\,\nabla_\theta \log \rho_\theta(q)
&= \sum_q \nabla_\theta \rho_\theta(q) \\
&= \nabla_\theta\!\left(\sum_q \rho_\theta(q)\right) \\
&= \nabla_\theta(1) = 0,
\end{split}
\end{equation}
where the third equality uses the fact that $\rho_\theta$ is a probability distribution and therefore sums to one identically in $\theta$.

\textbf{First sum simplifies via the log-derivative trick.} Applying $\nabla_\theta \rho_\theta(q) = \rho_\theta(q)\,\nabla_\theta \log \rho_\theta(q)$ and writing $\ell_\theta(q) := \log \rho_\theta(q)$ and $\ell_{\theta_0}(q) := \log \rho_{\theta_0}(q)$, we obtain the closed-form gradient
\begin{equation}
\label{eq:app-decoupling-final}
\begin{split}
\nabla_\theta &\mathcal{R}_{\mathrm{query}}(\theta) \\
&= \mathbb{E}_{q\sim \rho_\theta}\!\Big[\bigl(\ell_\theta(q) - \ell_{\theta_0}(q)\bigr) \\
&\qquad \cdot \nabla_\theta \ell_\theta(q)\Big].
\end{split}
\end{equation}

\textbf{Structural conclusion.} The right-hand side of Eq.~\eqref{eq:app-decoupling-final} depends on $\theta$ only through the query log-likelihood $\ell_\theta(q)$ (and its gradient $\nabla_\theta \ell_\theta(q)$); no factor of the response policy score function $\nabla_\theta \log \pi_\theta(o\mid q)$ appears. Equivalently, perturbations to $\pi_\theta(o\mid q)$ that leave the query marginal $\rho_\theta$ unchanged incur zero QKL gradient and are unconstrained by $\mathcal{R}_{\mathrm{query}}$. This establishes the structural decoupling between input-environment regularization (carried by QKL) and response-side exploration (left untouched). $\hfill\square$
    

%% file: tex/wq_append.tex
\section{Derivation of Query Reweighting}
\label{app:query-reweighting-derivation}

The standard RL objective optimizes the expected reward under the empirical training query distribution
\(\rho_{\mathrm{train}}\). ERPO instead considers a reference-aligned objective in which queries are sampled from the cached reference-induced query prior
\(\rho_{\theta_0}\):
\begin{equation}
\label{eq:app-erpo-objective-rhotheta0}
J_{\mathrm{ERPO}}(\theta)
=
\mathbb{E}_{q\sim\rho_{\theta_0},\,o\sim\pi_\theta(\cdot\mid q)}
\big[g(q,o)\big].
\end{equation}
Expanding the expectation gives
\begin{equation}
J_{\mathrm{ERPO}}(\theta)
=
\sum_q \rho_{\theta_0}(q)
\sum_o \pi_\theta(o\mid q)g(q,o).
\end{equation}
However, stochastic training samples queries from the empirical distribution
\(\rho_{\mathrm{train}}\), rather than directly from
\(\rho_{\theta_0}\). To express the same objective under
\(\rho_{\mathrm{train}}\), we multiply and divide by
\(\rho_{\mathrm{train}}(q)\):
\begin{align}
J_{\mathrm{ERPO}}(\theta)
&=
\sum_q
\rho_{\mathrm{train}}(q)
\frac{\rho_{\theta_0}(q)}{\rho_{\mathrm{train}}(q)}
\nonumber \\
&\qquad
\times
\sum_o \pi_\theta(o\mid q)\,g(q,o)
\nonumber \\
&=
\mathbb{E}_{q\sim\rho_{\mathrm{train}},\,o\sim\pi_\theta(\cdot\mid q)}
\left[
w^\star(q)g(q,o)
\right],
\end{align}
where the ideal importance weight is
\begin{equation}
\label{eq:app-ideal-query-weight}
w^\star(q)
=
\frac{\rho_{\theta_0}(q)}{\rho_{\mathrm{train}}(q)}.
\end{equation}

If training queries are sampled approximately uniformly from a dataset of size \(N\), then
\(\rho_{\mathrm{train}}(q)=1/N\). Therefore,
\begin{equation}
w^\star(q)
=
\frac{\rho_{\theta_0}(q)}{\rho_{\mathrm{train}}(q)}
=
N\rho_{\theta_0}(q).
\end{equation}
Since the multiplicative constant \(N\) is shared by all queries, it does not affect the relative weighting among queries. Thus, up to a global constant,
\begin{equation}
w^\star(q)
\propto
\rho_{\theta_0}(q).
\end{equation}

In practice, ERPO estimates the reference-induced query prior using a cached reference score
\(\ell_{\theta_0}(q)\), which assigns larger values to queries that are more favored under the pretrained reference model. Hence we use a bounded, dataset-static query weight \(w(q)\) that is monotone in
\(\ell_{\theta_0}(q)\):
\begin{equation}
w(q)
\propto
\ell_{\theta_0}(q).
\end{equation}
Substituting this query weight into the negative optimization objective yields the ERPO training loss
\begin{align}
\label{eq:app-erpo-loss-query-weighted}
L_{\mathrm{ERPO}}(\theta)
&=
-
\mathbb{E}_{q\sim\rho_{\mathrm{train}},\,o\sim\pi_\theta(\cdot\mid q)}
\left[
w(q)g(q,o)
\right] \nonumber \\
&\quad
+
\alpha\mathcal{R}_{\mathrm{query}}(\theta).
\end{align}
where \(\mathcal{R}_{\mathrm{query}}(\theta)\) denotes the query-level regularization term and \(\alpha\) controls its strength.

In practice, the exact reference-induced query prior
\(\rho_{\theta_0}(q)\) is not directly available on a finite training set.
Therefore, we use the cached reference-model log-probability as a practical monotone proxy for the relative query preference induced by the reference model.
This approximation only aims to preserve the relative ordering of queries under the reference model, rather than to estimate the absolute density value.
We emphasize that this implementation is not intended as an unbiased density-ratio estimator; rather, it is a bounded monotone approximation to the ideal query importance weight in Eq.~\eqref{eq:app-ideal-query-weight}.

For each query \(q_i\), let
\(\log p_{\theta_0}(q_i)\) denote its cached sequence log-probability under the reference model.
We first define its negative log-probability as
\begin{equation}
s_i
=
-\log p_{\theta_0}(q_i).
\end{equation}
A larger reference-model probability corresponds to a smaller \(s_i\).
To obtain a weight whose magnitude increases with the reference-model query probability, we use the inverse normalized negative log-probability.
For a dataset with \(N\) queries, let
\begin{equation}
\bar{s}
=
\frac{1}{N}
\sum_{j=1}^{N}
s_j.
\end{equation}
We define the unbounded query weight as
\begin{equation}
\tilde{w}(q_i)
=
\frac{\bar{s}}{s_i}.
\end{equation}
This normalization makes the weight dimensionless and centers its scale around the dataset average, while preserving the monotonic relationship that higher-probability queries receive larger weights.
Finally, to reduce the variance of importance reweighting and prevent any single query from dominating optimization, we clip the weight to a fixed range:
\begin{equation}
w(q_i)
=
\mathrm{clip}
\left(
\tilde{w}(q_i), 0, 2
\right).
\end{equation}